# Delta Learning Rule for the Active Sites Model

Krishna Chaithanya Lingashetty

*Abstract : This paper reports the results on methods of comparing the memory retrieval capacity of the Hebbian neural network which implements the B-Matrix approach, by using the Widrow-Hoff rule of learning. We then, extend the recently proposed Active Sites model by developing a delta rule to increase memory capacity. Also, this paper extends the binary neural network to a multi-level (non-binary) neural network.*

**Introduction**

The motivation for studying neural networks comes from the fact that computers have been unable to generate intelligent behavior similar to that of biological organisms. How does a biological neural network process the information that it receives? How can it remember and retrieve memories? The attempt to understand how a biological neural network works has been made from several perspectives including flow of information from primary centers [1]-[3], resonance in feedback circuits [4], and the capacity of neural networks [5]-[11]. There is also higher level analysis of neural network models compared to other computational models [12]-[26]. Specific issues related to human memory are considered in [14],[23]-[26].

We assume a hierarchy amongst neurons and that specific parts of this hierarchy are to carry out several different physiological functions. As mentioned in an earlier paper [2], the neural network might be made up of a several indexing neural nets, which sit on top of sub-neural networks. Stimulation of indexing neural networks, would then in turn cause a stimulation in its lower hierarchy until the signal reaches a sub-neural network. Once this signal activates a neuron or a set of neurons in the sub-neural net then, the activity spreads from them to the neighboring neurons until a memory stored in that network is successfully retrieved. This was the idea that was proposed by Kak [1],[3] in previous papers. The B-Matrix approach [1] provides a way of retrieving memories from the network by activity spreading.

Recollection of memories in the B-Matrix Approach is by using the lower triangular matrix *B*, constructed as,

$$T = B + B^t$$

Starting with the fragment $f_1$, the updating proceeds as:

$$f_i = \text{sgn}(B \cdot f_{i-1})$$

where $f_i$ is the $i^{th}$ iteration of the generator model. Notice that the $i^{th}$ iteration of the generator model produces only the value of the $i^{th}$ binary index of the vector memory but does not alter the "i-1" values already present. The assumption is that this signal that is given to a particular neural network would be directed at a particular set of neurons which give us a better chance at retrieving memories from the network.

One way to extend the capacity of a neural network is by the use of non-binary networks [9]. We propose a way to do so in this paper for the B-matrix approach. We also show how the Widrow-Hoff learning rule may be applied to the B-Matrix approach for memory retrieval.

**Widrow-Hoff Learning:**

The Widrow-Hoff learning rule was proposed to increase the memory storage capacity of the Hebbian network. In the Widrow-Hoff model, we try to adjust the weights that are stored in the network iteratively to increase the chances of retrieving memories from the network. the idea behind this mode of learning is that as new memories are brought into the network, the learning of these new memories would have an overwriting effect on the previously learned memory.

The way that the Widrow-Hoff model proceeds is that it first tries to calculate the error associated with trying to retrieve the memory from the network. Then based on the error matrix obtained, the weights are thus adjusted such that the error for the particular memory is minimized. This procedure is repeated iteratively until all the memories of the network have been stored with no error, or with a permissible threshold. As is well known, with Widrow-Hoff learning, the connection weight between two neurons is iteratively adjusted in order to have a smaller error term for a second presentation of the same stimulus.

$$W_{n+1} = W_n + \Delta(W_n)\text{ , where }\Delta(W_n) = \eta(x_i - W_n x_i),$$ W is the weight matrix, $x_i$ is the present input, and $\eta$ is a small positive constant.

This way of learning is usually called as "batch learning" as opposed to "single stimulus learning" as proposed in Hebbian Learning. It has been shown that batch learning does converge faster to the correct solution than single stimulus learning.

There is an error vector that is estimated for every iteration of the weight matrix adjustment. We then calculate an error term associated with these vectors, and average it over the number of memories that are trained to the network. So defining a good bound on this error term forms a critical problem.

**Multi-Level/Non-Binary Neurons**

All the experiments discussed until now, were considering binary neural networks. The problem with using such a network is that these are no comparison to the neural networks that are prevalent in biological organisms. The biological neurons are subject to not only electrical impulses of varying magnitude, but also a number of such spikes called the spike train. Conversions of all the data that is being fed to the network to binary can, in actuality, cause loss of information. Hence, non-binary or n-ary neural networks have been proposed [9] taking into consideration, the complexity of biological neural networks.

Consider the construction of a quaternary neural network instead of a binary neural network. The quaternary neural network [9] implements the same principles as does a binary, with the exception that the neurons now, map to a larger set of

$$x_i = \sum_j T_{i,j} V_j$$

$$V_i = \begin{cases} -3 & x_i < -t \\ -1 & x_i < 0 \\ 1 & x_i < t \\ 3 & x_i > t \end{cases}$$

The problem involved in extending this approach to the B-Matrix approach would involve the inclusion of varying levels of thresholds at each step of the multiplication of the B-Matrix to the fragment, considering that there are a different number of neurons involved in each step. Specifically, the number of neurons involved increases by 'one' in each step. Hence, the selection of a good function which would accommodate the number of neurons as they get incremented in each step forms a highly improbable task..

**Active Sites Model**

The Active sites model gives a fresh insight as to how memories can be indexed and retrieved in a neural network with very less computational complexity involved [2]. The idea behind the Active sites model comes from the fact that every memory vector being trained to the network has a unique fragment of memory as compared to the other memories that are being trained on the network. The neuron sites, in which this unique fragment of memory resides, become the active sites for a particular memory. These sites, when triggered with the right input stimulus, might give us a better chance at retrieving memories than the regular approach.

In this model, a network can potentially identify a neuron as an active site by assigning a n activation level for it while adjusting the synaptic strength between two excited neurons. The Active sites model has been quite effective in reducing the computation required for retrieving memories and has also increased the sustained retrieval of memories through the neural network.

**Delta Rule**

The Widrow-Hoff learing rule was implemented with one goal in mind, to incrementally increase the memory retrieval capacity of the Hebbian model. At each iteration of a particular memory, the effectiveness of the retrieval capacity of the network is improved. Hence, comparing the results of the increase in Hebbian model to the B-Matrix approach or the Active sites model would not be justified.

Since at each step, the delta rule computes the change it needs to make to the weight matrix such that it can accommodate the incoming memory into the network. For the same result to be expected from a delta rule for the B-Matrix approach, we need to know which site has to be selected for learning the particular memory. Hence, the Active sites model becomes the default model for implementing a delta rule for the B-Matrix approach.

Once the site/sites are chosen by the Active sites model, the delta rule would then update the rows individually until the desired memory is retrieved from the weight matrix.

$$\Delta_{i+1,k} = 0, \quad \text{if } sgn(B.f_i) = input_{i+1}$$

$$= sgn(input_{i+1} - sgn(B.f_i)) * input_k * \eta, \quad otherwise$$

$$where\ \eta \text{ is the learning constant}$$

The delta rule discussed can be applied to a non-binary neural network, as we can specify the threshold individually for each level at learning. Hence this model solves the problem of implementing a non-binary neural network for a B-Matrix approach.

The delta rule can be implemented with having one Active site per memory and multiple active sites per memory. Forcing the restriction of having only one active site per memory, gives us only 16 unique active sites per memory and hence, a maximum of only 16 memories can be trained to the network. On removing that restriction however, we have many combinations of active sites available and hence, more probability to store memories. For generating the update order of multiple active sites, the averaged method mentioned in [2] is used.

**Results** The following graphs present the results of our simulations.

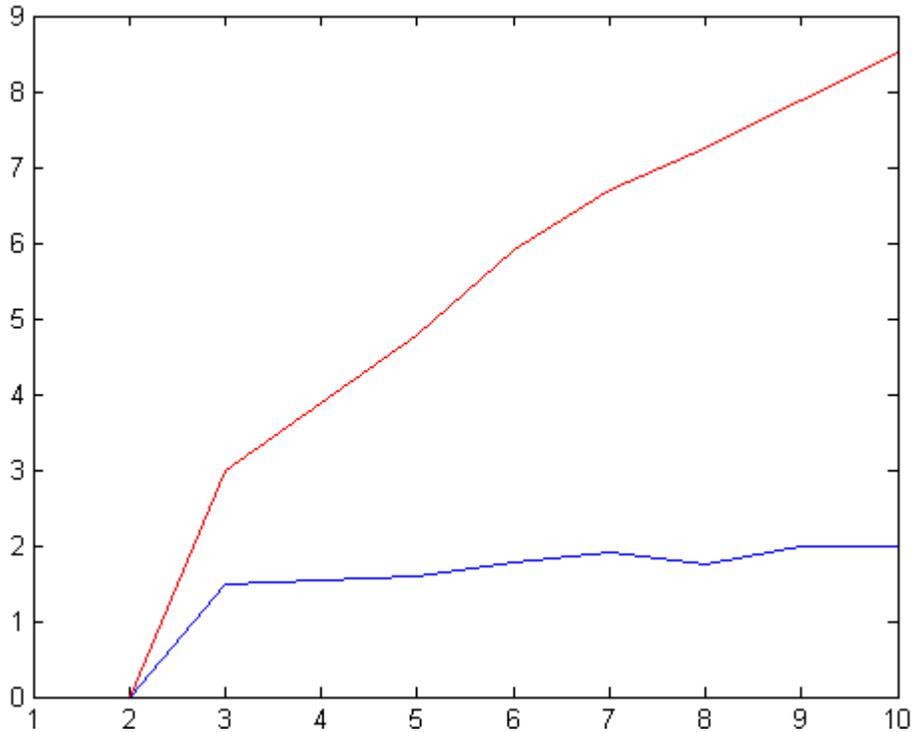

Fig. 1, 12 Neurons, Active sites model with the Widrow-Hoff learning rule

In fig. 1, we notice that the number of memories retrieved never goes higher than 2 memories, no matter how many memories are fed to the network. Hence, using the Widrow-Hoff learning rule to the Active sites/B-Matrix approach would not be justifiable as this rule was developed to increase memory capacity of a Hebbian model but not the B-Matrix approach as such.

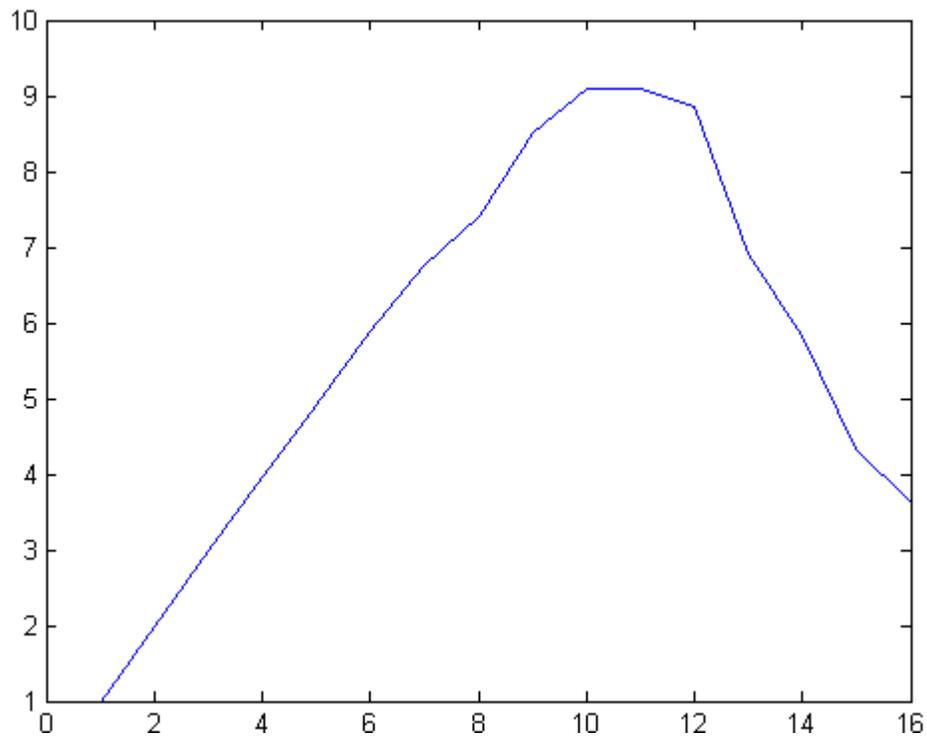

Fig. 2, Delta Rule for AS with 16 neurons

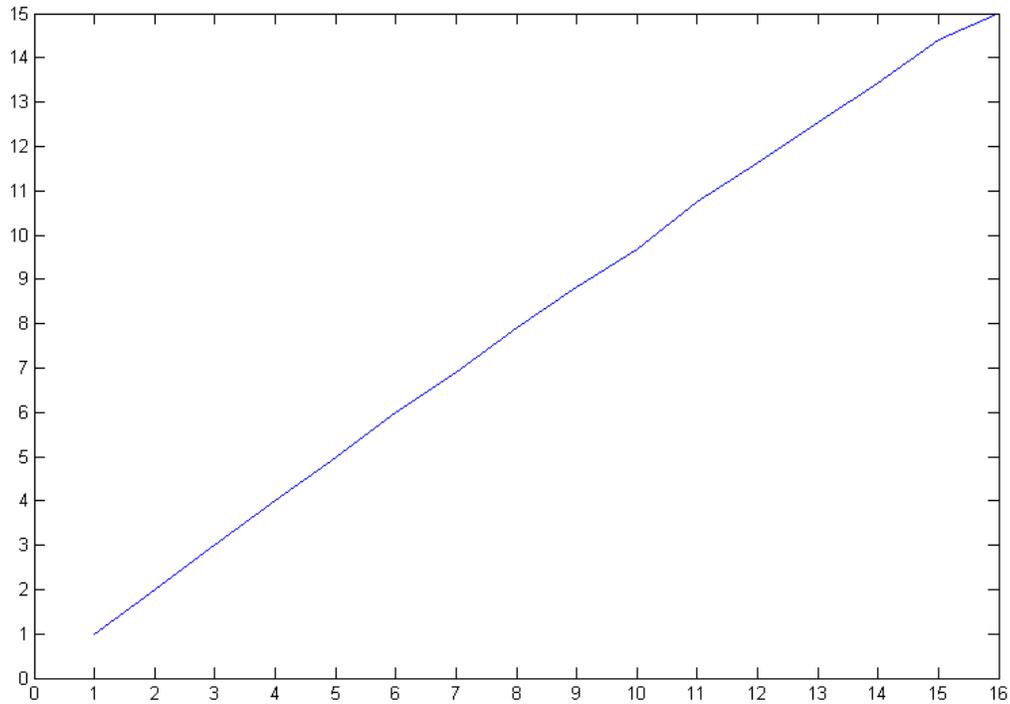

Fig. 3, Delta Rule for AS with 32 neurons

Fig. 2 and 3 above show the results of the delta rule applied to the Active sites model. We can notice from the figures that the number of memories that are being retrieved from the neural network have increased significantly. The increase in retrieval capacity of the neural network has increased more than a 100% as compared to the Active sites approach. The number of retrieved memories peaks off slightly over the n/2 mark, where n is the number of neurons in the network. For a neural network of size 16, we can retrieve nearly 9 memories out of the 10 that are trained to the network. After that, the number of memories stored falls down. The reason for this could be that since 16 neurons are available for potential active sites, the later memories are given active sites where they aren't actually as active, but they keep rewriting the already learnt memories, which reduces retrieval.

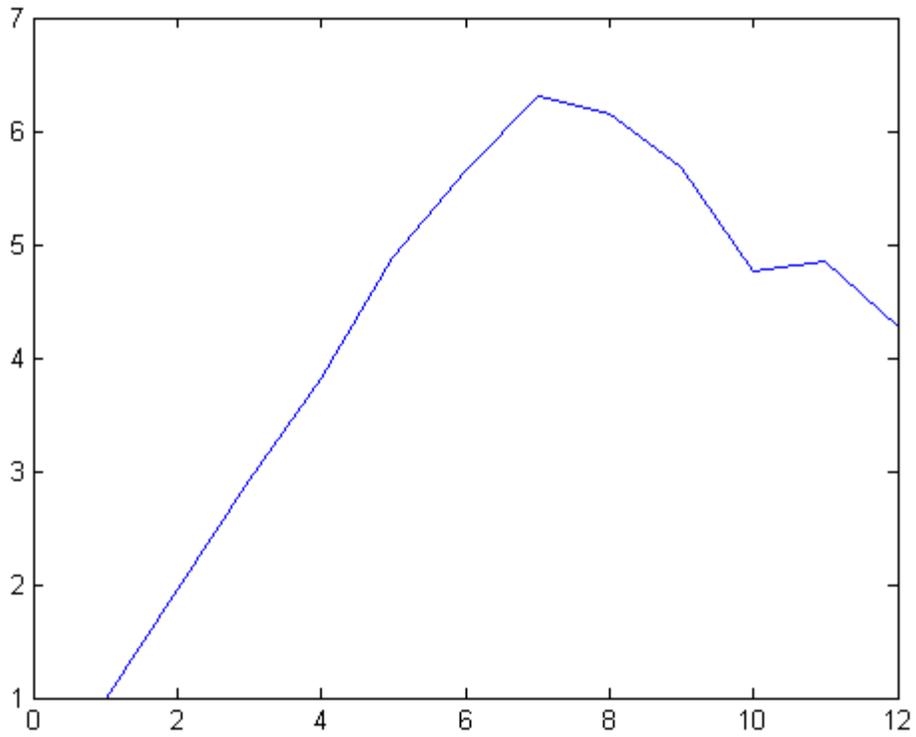

Fig. 4, Non-Binary Delta Rule with 16 neurons and one active site per memory

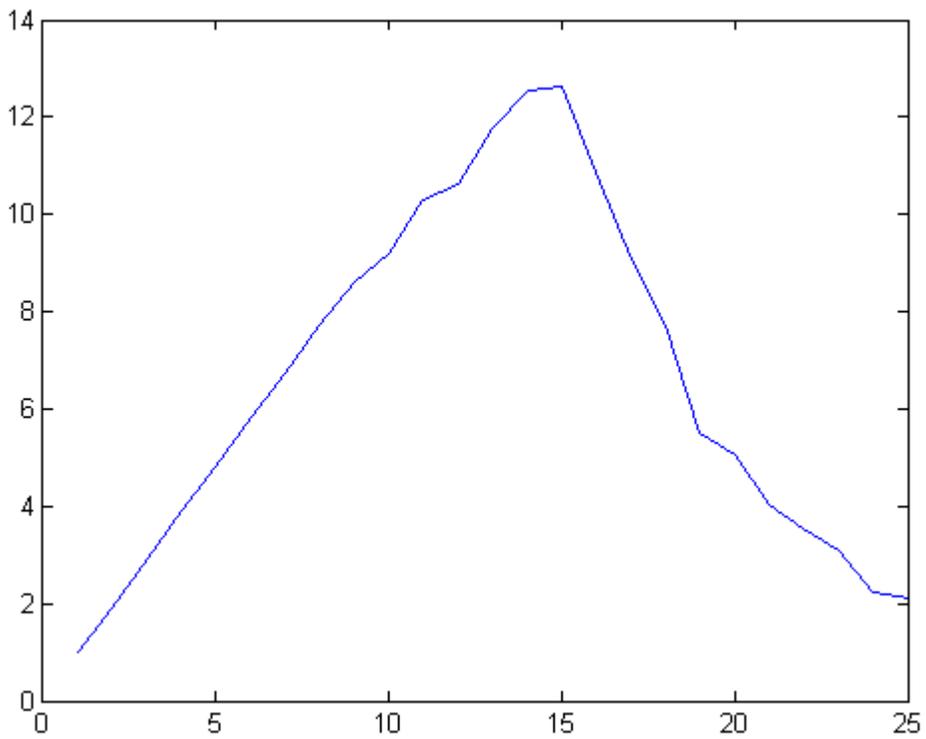

Fig. 5, Non-Binary Delta Rule with 32 neurons

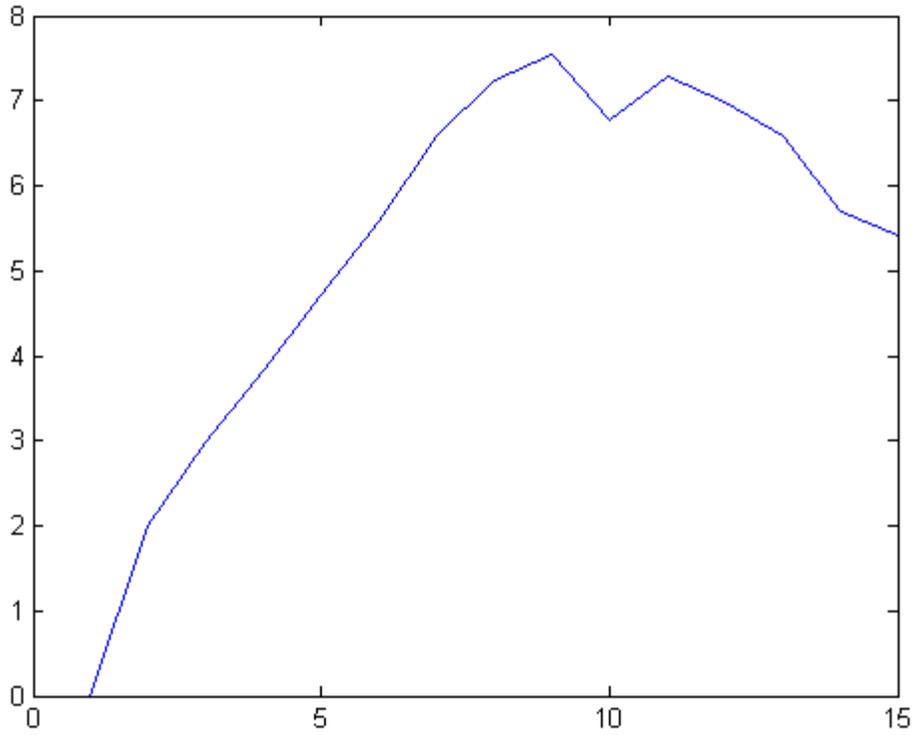

Fig. 6, Non-Binary Delta Rule with 16 neurons and 2 active sites per memory

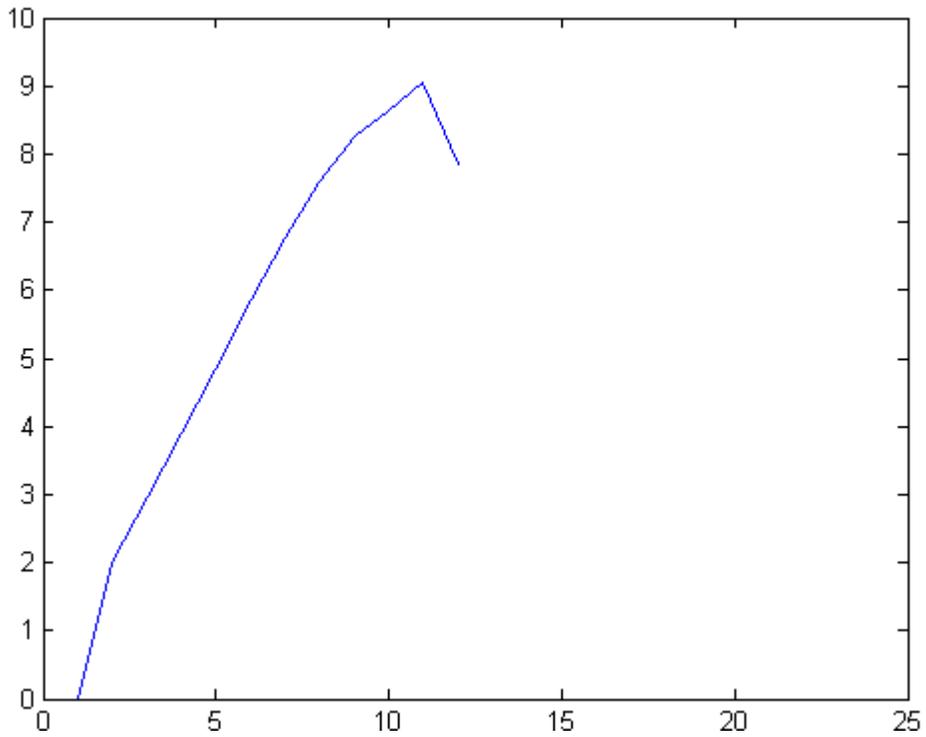

Fig. 7, Non-Binary Delta Rule with 16 neurons and 3 active sites per memory

Figures 4 and 5 show the delta rule being applied to a non-binary neural network to networks of size 16 and 32 respectively. We can notice that the non-binary retrieval does not fare as well as the binary neural network. The maximum retrieved memories peaks off at around 6.5 and 13 for the non-binary model as compared to the 9 and 16 of the binary model. However, considering the amount of information that is stored in these networks, the non-binary memory incorporates four times as much information than the binary model. Hence, the non-binary model stores more information than the binary model.

Figures 6 and 7 show the application of the delta rule with more than one active sites per memory. As we can notice, the number of memories retrieved successfully increases with the increase in the number of active sites(6.5 for one active site, 7.5 for 2 active sites and 9 for three active sites). Increasing the number of active sites does increase memory retrieval but increases the computational complexity of the model, as there are an increased number of combinations of sites and the fragments of memory.

**Conclusion**

In conclusion, the Widrow-Hoff model proposed for higher retrieval capacity of the Hebbian model does not increase the memory retrieval capacity of the B-Matrix or the Active Sites model. The proposed delta rule increases the memory retrieval capacity of the neural network by more than a 100% and using more active sites per memory increases the retrieval capacity.

The delta rule thus proposed gives us a different dimension in perceiving how a complex biological neural network might perform. A complex biological network need not be a binary model, instead the communication between the neurons and the brain has always been thought of to be a train of electro-chemical signals. Hence the introduction of the non-binary neural networks provides an insight into the possible way of storing and retrieving more information than traditional binary neural networks.

# References

[1] S. Kak, Feedback neural networks: new characteristics and a generalization. Circuits, Systems, and Signal Processing, vol. 12, pp. 263-278, 1993.
[2] K.C. Lingashetty, Active sites model for the B-matrix approach. 2010. arXiv:1006.4754
[3] S. Kak, Single neuron memories and the network's proximity matrix. 2009. arXiv:0906.0798
[4] J.J. Hopfield, Neural networks and physical systems with emergent collective computational properties. Proc. Nat. Acad. Sci. (USA), vol. 79, pp. 2554-2558, 1982.
[5] D.L. Prados and S. Kak, Neural network capacity using the delta rule. Electronics Letters, vol. 25, pp. 197-199, 1989.
[6] S. Kak, The three languages of the brain: quantum, reorganizational, and associative. In: K. Pribram, J. King (Eds.), Learning as Self- Organization, Lawrence Erlbaum, London, 1996, pp. 185-219.
[7] R. J. McEliece, E. C. Posner, E. R. Rodemich, and S. S. Venkatesh, The capacity of the Hopfield associative memory, IEEE Trans. Inform.Theory, vol. IT-33, pp. 461–482, 1987.
[8] K.H. Pribram and J. L. King (eds.), Learning as Self-Organization. Mahwah, N. J.: L. Erlbaum Associates, 1996.
[9] D. Prados and S. Kak, Non-binary neural networks. Lecture Notes in Computing and Control, vol. 130, pp. 97-104, 1989.
[10] M. Schuster and K. K. Paliwal. Bidirectional recurrent neural networks. IEEE Transactions on Signal Processing, vol. 45, pp. 2673–2681, 1997.
[11] C. Ji and D. Psaltis, Capacity of two-layer feedforward neural networks with binary weights. IEEE Trans. Inform. Theory, vol. 44, pp. 256-268, 1998.
[12] S. Kak, Can we define levels of artificial intelligence? Journal of Intelligent Systems, vol. 6, pp.133-144, 1996.
[13] S. Kak, Artificial and biological intelligence. ACM Ubiquity, vol. 6, number 42, pp. 1-20, 2005.
[14] D.L. Schacter, Searching for Memory: The Brain, the Mind, and the Past. Basic Books, New York, 1997.
[15] G.D.A. Brown, I. Neath, N. Chater, A ratio model of scale-invariant memory and identification. Psychological Review, vol. 114, pp. 539-576, 2007.
[16] S. Kak, New training algorithm in feedforward neural networks, First International Conference on Fuzzy Theory and Technology, Durham, N. C., October 1992. Also in Wang, P.P. (Editor), Advance in fuzzy theory and technologies, Durham, N. C. Bookwright Press, 1993.
[17] S. Kak, A class of instantaneously trained neural networks, Information Sciences, 148, 97-102, 2002.
[18] S. Kak, On generalization by neural networks. Information Sciences, vol. 111, pp. 293-302, 1998.
[19] S.Kak, New algorithms for training feedforward neural networks. Pattern Recognition Letters, 15, 295-298, 1994.
[20] M.C. Stinson and S. Kak, Bicameral neural computing. Lecture Notes in Computing and Control, vol. 130, pp. 85-96, 1989.
[21] S. Kak, Multilayered array computing, Information Sciences, vol. 45, pp. 347-365, 1988.
[22] S. Kak and J. F. Pastor, Neural networks and methods for training neural networks. US Patent 5,426,721, 1995.
[23] N. Cowan, Working Memory Capacity. Psychology Press, New York, 2005.

[24] S. Kak, Active agents, intelligence, and quantum computing, Information Sciences, vol. 128, pp. 1-17, 2000.
[25] U. Neisser and I. Hyman, Memory Observed. Worth Publishing, 1999.
[26] D.L. Schacter, The Seven Sins of Memory: How the Mind Forgets and Remembers. Mariner Books, 2002.